\documentclass[conference]{IEEEtran}
\IEEEoverridecommandlockouts
\usepackage{cite}
\usepackage{amsmath,amssymb,amsfonts}
\usepackage{algorithmic}
\usepackage{graphicx}
\usepackage{textcomp}
\usepackage{xcolor}
\usepackage{subcaption}

\def\BibTeX{{\rm B\kern-.05em{\sc i\kern-.025em b}\kern-.08em
    T\kern-.1667em\lower.7ex\hbox{E}\kern-.125emX}}
\begin{document}

\title{MonoSparse-CAM: Efficient Tree Model Processing via Monotonicity and Sparsity in CAMs\\
\thanks{This work is supported by the National Science Foundation (NSF) under grants IIS-2332744, ECCS-2328712, CCF-2328805, and CNS-2112562. The views, opinions, and/or findings contained in this article are those of the authors and should not be interpreted as representing the official views or policies, expressed or implied, by the NSF.
}
}


\author{\IEEEauthorblockN{Tergel Molom-Ochir,
Brady Taylor, Hai (Helen) Li and Yiran Chen}
\IEEEauthorblockA{Department of Electrical and Computer Engineering,
\\Duke University, Durham, North Carolina\\
Email: \{tergel.molom-ochir,
brady.g.taylor,
hai.li,
yiran.chen\}@duke.edu}}


\maketitle

\begin{abstract}

While the tree-based machine learning (TBML) models exhibit superior performance compared to neural networks on tabular data and hold promise for energy-efficient acceleration using aCAM arrays, their ideal deployment on hardware with explicit exploitation of TBML structure and aCAM circuitry remains a challenging task. In this work, we present MonoSparse-CAM, a new CAM-based optimization technique that exploits TBML sparsity and monotonicity in CAM circuitry to further advance processing performance. Our results indicate that MonoSparse-CAM reduces energy consumption by upto to 28.56$\times$ compared to raw processing and by 18.51$\times$ compared to state-of-the-art techniques, while improving the efficiency of computation by at least 1.68$\times$.


\end{abstract}


\section{Introduction}
Artificial intelligence (AI) has radically impacted many fields, such as computer vision, robotics, and healthcare \cite{AIintro1, AIintro2, AIintro3}, in which deep neural networks play a crucial role. However, TBML often yields higher performance than deep learning models when applied to tabular data, apart from enjoying advantages of interpretability and debug ease \cite{TBML>DL1, TBML>DL2, TBML>DL3}. Therefore, it becomes highly relevant to explore the hardware implementation of TBML. One of the more well-known approaches is the mapping of TBML onto analog content-addressable memory (aCAM) arrays \cite{b2}.

The literature provides a comparison of TBML accelerators, highlighting that memristive analog CAMs correspond to the lowest energy consumption option \cite{b2, energypercell}. In fact, architecture optimization is a crucial task with regard to TBML accelerators for efficiently processing large arrays, which are difficult to fabricate. Deployment management was also considered in previous works for increasing performance in processing. Feature Reordering (FR) is a  methodology in the architecture optimization area and relies on reordering rows and columns according to activity, bringing the most active cells to the bottom left corner of the array and processing them\cite{b2}. In raw processing, tasks are performed without any energy-saving or performance-enhancing techniques. FR can save time and energy by prioritizing active cells while skipping the empty ones. It provides a slight improvement in throughput. In the case of CAMs, different matching techniques can be employed for power efficiency; while many techniques employ the use of the exact matching technique in CAMs, other potentially power-efficient matching criteria like best match, threshold match, and partial match have not yet been considered in hardware implementation \cite{b6}.


This paper proposes and discusses MonoSparse-CAM, a novel software-hardware co-design for accelerating TBML models on CAM arrays. MonoSparse-CAM leverages both the sparsity in the decision trees and monotonicity in the CAM array to optimize computation. Key contributions include:
\begin{itemize} 
  \item Introducing MonoSparse-CAM, a CAM-based optimization algorithm that exploits TBML sparsity and array monotonicity. 
  \item Analyzing how tree-based model structures influence array sparsity, emphasizing the need for tree-aware optimization. 
  \item Demonstrating that MonoSparse-CAM addresses scalability issues in large arrays and significantly reduces CAM energy consumption.
\end{itemize} 
Our results showed that structural balance in TBML models is a major determinant of array sparsity and acts as an entry point for further optimizations. Regardless of the sparsity level, MonoSparse-CAM minimizes energy consumption with high accuracy and throughput. MonoSparse-CAM achieves 3.45$\times$ less energy consumption on highly sparse arrays and 18.51$\times$ on low-sparsity arrays compared to the state-of-the-art techniques.

\section{Background}

\begin{figure}[b]
    \begin{minipage}{0.5\textwidth}%
      \begin{subfigure}[t]{0.4\linewidth}
        \centering
        \includegraphics[width=\linewidth]{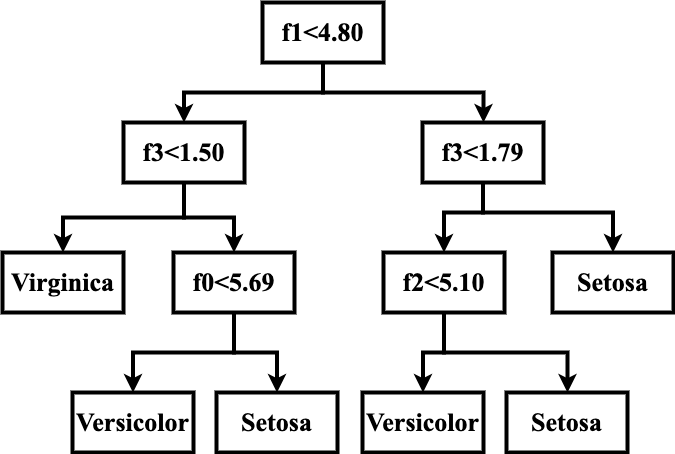}
        \caption{}
        \label{subfig:example_decision_tree}
      \end{subfigure}
      \begin{subfigure}[t]{0.55\linewidth}
        \centering
        \includegraphics[width=\linewidth]{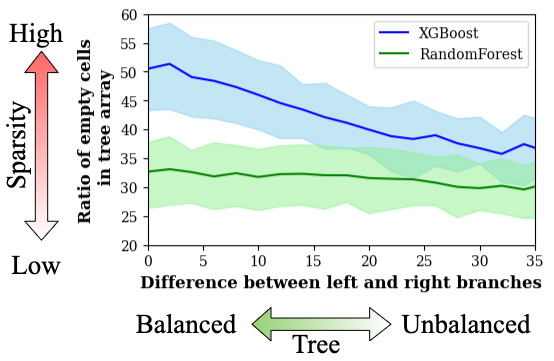}
        \caption{}
        \label{subfig:trend_in_DT}
      \end{subfigure}
    \end{minipage}%
    \caption{(a) A 4-feature XGBoost decision tree trained on the Iris Dataset \cite{iris_dataset}. (b) As shown in Section \ref{3.2}, experiments indicate that as binary trees become more balanced, they also become sparser. Further details are provided in Section \ref{3.2}.}
\end{figure}

Tree-based models, such as decision trees, Random Forest, and XGBoost, are widely used for classifying tabular data by sequentially comparing feature values at each node until reaching a leaf node as shown in \ref{subfig:example_decision_tree}. These models remain state-of-the-art for tabular data due to their interpretability, cost-effectiveness, and robust performance \cite{DL_Tabular}. Although deep neural networks (DNNs) are gaining popularity in this domain, TBML models consistently outperform them in both traditional and federated learning settings, offering higher accuracy, lower computational cost, and reduced carbon footprint \cite{TBML>DL1, TBML>DL2, TBML>DL3, TBML>DL4, GreenAI, TBML>DL5}. The inherent sparsity and balance of tree-based models significantly impact their performance, with certain tree structures being more suitable for specific optimization algorithms. However, methods like FR are less effective for denser trees, further motivating an analysis of how tree architecture influences TBML processing efficiency.

Analog CAM arrays efficiently search data by matching input voltages to stored ranges, using memristors in common-source amplifiers for threshold adjustment. These arrays have been employed to implement TBML models, where decision tree branch paths from source node to leaf nodes map to rows and data features to columns, with optimization techniques such as FR enhancing data compressibility. However, CAMs are known for high power consumption due to large-scale activation across the array \cite{high_power}. Recent innovations, such as hybrid CAM designs that combine the speed of NOR-type CAMs with the energy efficiency of NAND-type CAMs, and memristor-based TCAM circuits, aim to address this limitation \cite{inbook, hybrid_CAM}. Our method builds on these advancements to achieve both energy-efficient and high-speed processing within CAM-based architectures.

\begin{figure}[b]
    \begin{minipage}{0.5\textwidth}%
      \begin{subfigure}[t]{0.50\linewidth}
        \centering
        \includegraphics[width=\linewidth]{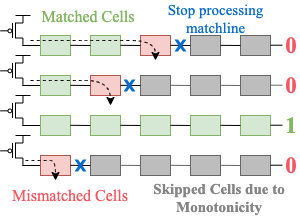}
        \caption{}
        \label{subfig:paperteaser2}
      \end{subfigure}
      \begin{subfigure}[t]{0.48\linewidth}
        \centering
        \includegraphics[width=0.95\linewidth]{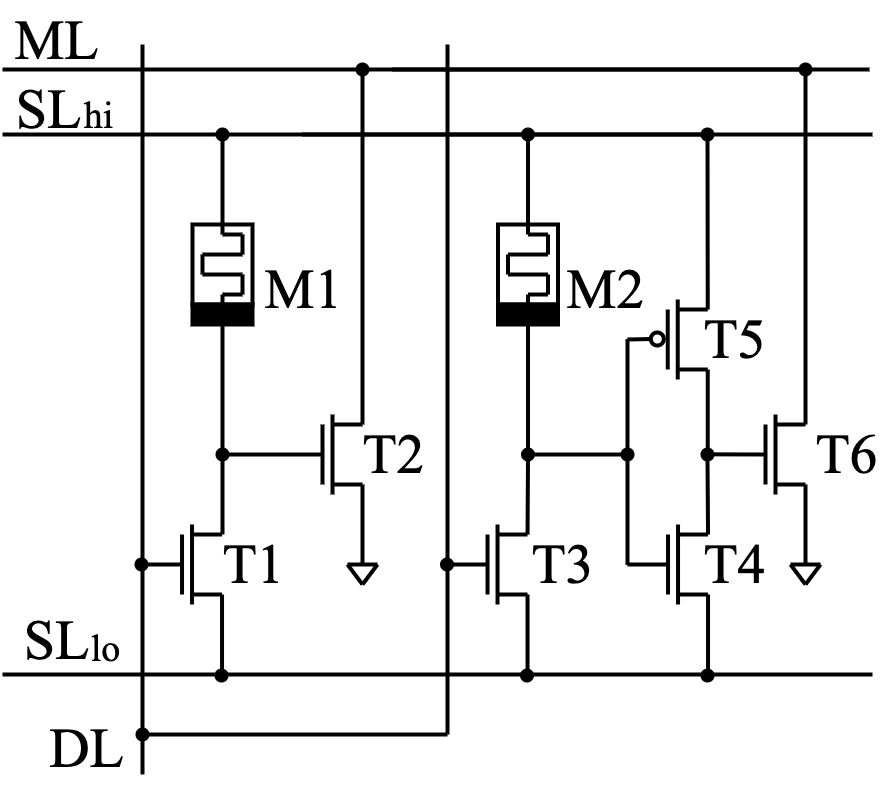}
        \caption{} 
        \label{subfig:cam_cell}
      \end{subfigure}
    \end{minipage}%
    \caption{(a) Illustration of the proposed technique: matched cells (green) continue processing, while mismatched cells (red) trigger early stops to save energy. Gray cells are skipped due to monotonicity. (b) The 6T2M analog CAM cell design, highlighting key components for matching operations, including transistors (T1-T6) and memristors (M1, M2).}
\end{figure}

\begin{figure*}[t]
  \centering
  \begin{subfigure}[t]{0.23\linewidth}
    \centering
    \includegraphics[width=\linewidth]{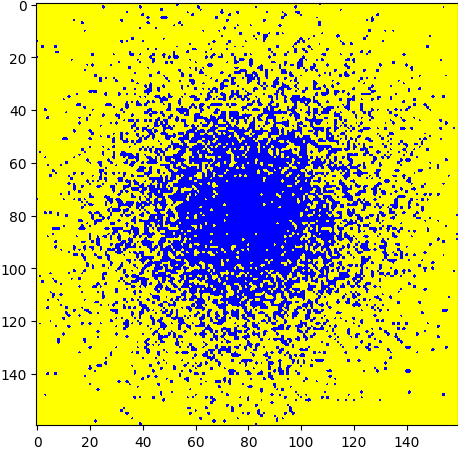}
    \caption{}
    \label{subfig:before_reodering}
  \end{subfigure}
  \begin{subfigure}[t]{0.23\linewidth}
    \centering
    \includegraphics[width=\linewidth]{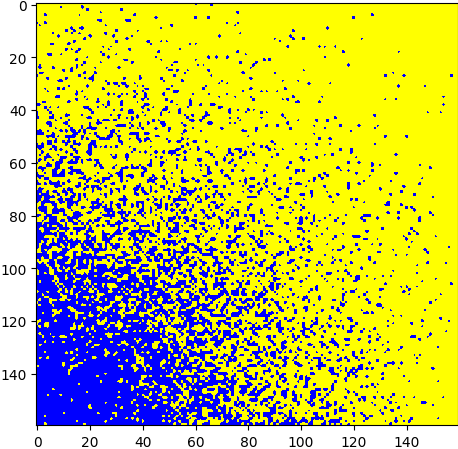}
    \caption{}
    \label{subfig:after_reordering}
  \end{subfigure}
  \begin{subfigure}[t]{0.23\linewidth}
    \centering
    \includegraphics[width=\linewidth]{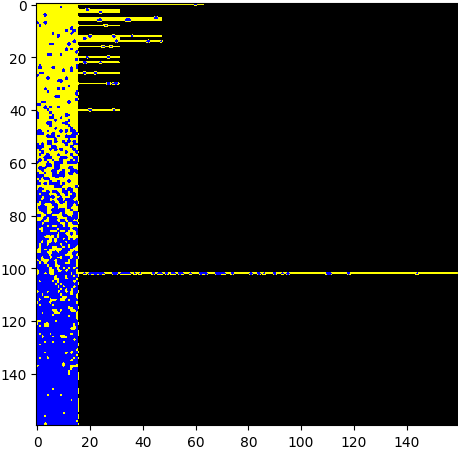}
    \caption{}
    \label{subfig:after_monosparse}
  \end{subfigure}

 \caption{
 (a) A 160$\times$160 decision tree grid with features and outcomes, simulated using a Gaussian distribution ($\lambda$ = 0.6, $\mu$ = 0.0). Yellow cells are inactive, blue cells are active. (b) The grid after Feature Reordering. (c) After applying MonoSparse-CAM, black cells show skipped computations, reducing energy use.}
 \label{subfig:blue_yellow_arrays}
\end{figure*}

\section{Methodology}
\subsection{MonoSparse-CAM Algorithm}

The MonoSparse-CAM algorithm improves FR by rearranging not only the active cells but also by bypassing deactivated rows throughout the whole processing stage. Cells with "don't care" values ("X") and deactivated cells, marked using a register, are not loaded to the CAM array. In a cycle, the active cells are processed, and the results of the match are stored in a register. In the following cycle, only the "alive"  matchlines are processed, hence saving unnecessary computation. Monotonicity is utilized in designing the matchline architecture of CAM. MonoSparse-CAM ensures that mismatched row is not taken into account as illustrated in Fig.\ref{subfig:paperteaser2}, thereby reducing unnecessary operations and energy consumption. Fig.\ref{subfig:blue_yellow_arrays} shows a sample array and how much of it has the potential to be skipped.


We generated the sparsity levels in these decision tree arrays by cell sampling using a Gaussian distribution. Let $\lambda$ be the sparsity control parameter, where higher values of $\lambda$ result in sparse arrays with a less number of active cells, and vice-versa. Hence, we vary $\lambda$ to simulate different real-world scenarios in which decision trees may contain both sparse and dense sets of features of interest. This guarantees that our simulations span the range of sparsity, from very sparse to much denser configurations that are computationally more demanding.



\subsection{Sparsity Considerations and Tradeoffs} \label{3.2}

\begin{figure*}[t]
  \centering
  \begin{minipage}{\textwidth}%
    \begin{minipage}{\textwidth}%
      \begin{subfigure}[t]{0.33\linewidth}
            \centering
        \includegraphics[width=\linewidth]{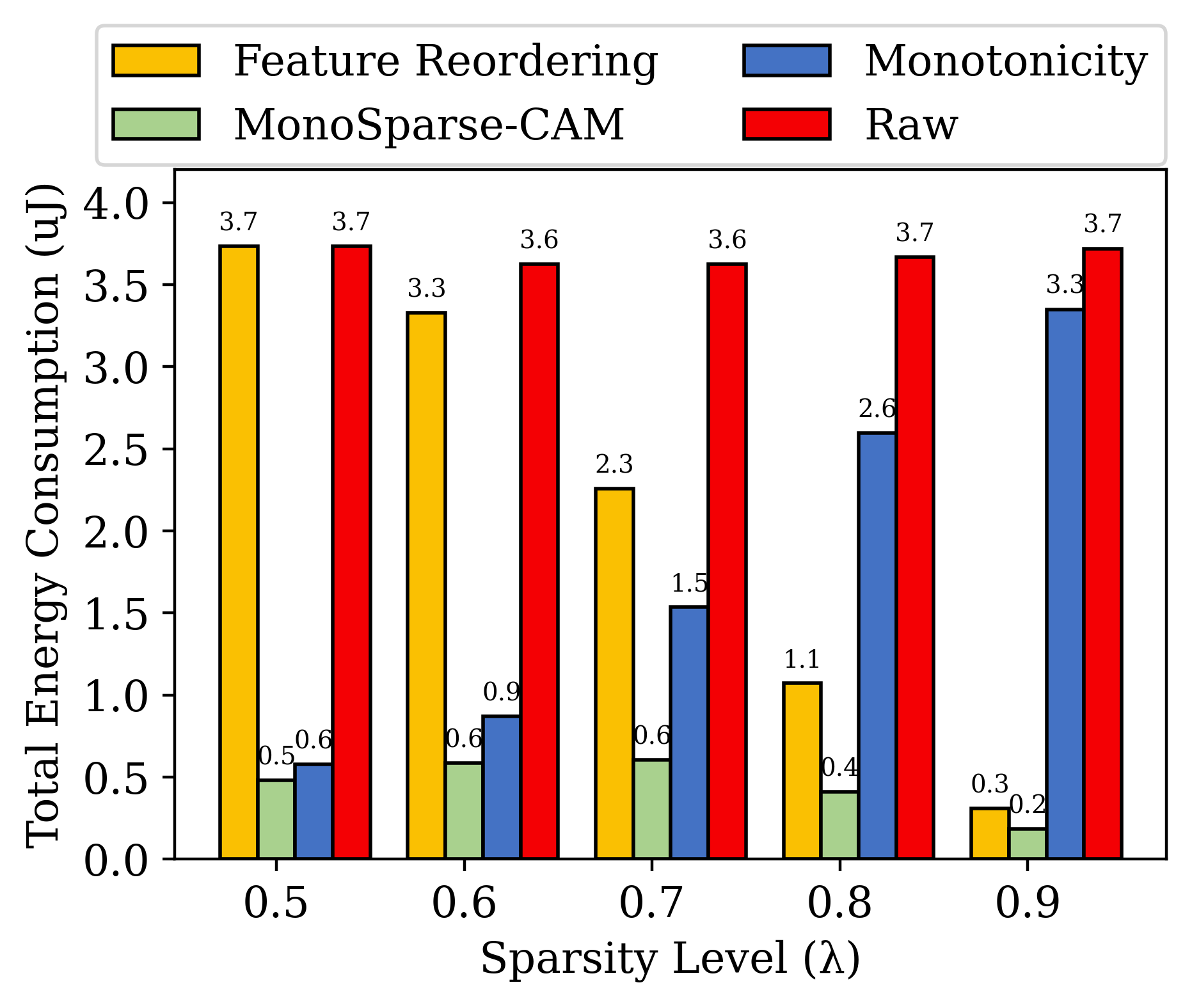}
        \caption{}
        \label{subfig:energy_consumption_comparison}
      \end{subfigure}
      \begin{subfigure}[t]{0.33\linewidth}
            \centering
        \includegraphics[width=\linewidth]{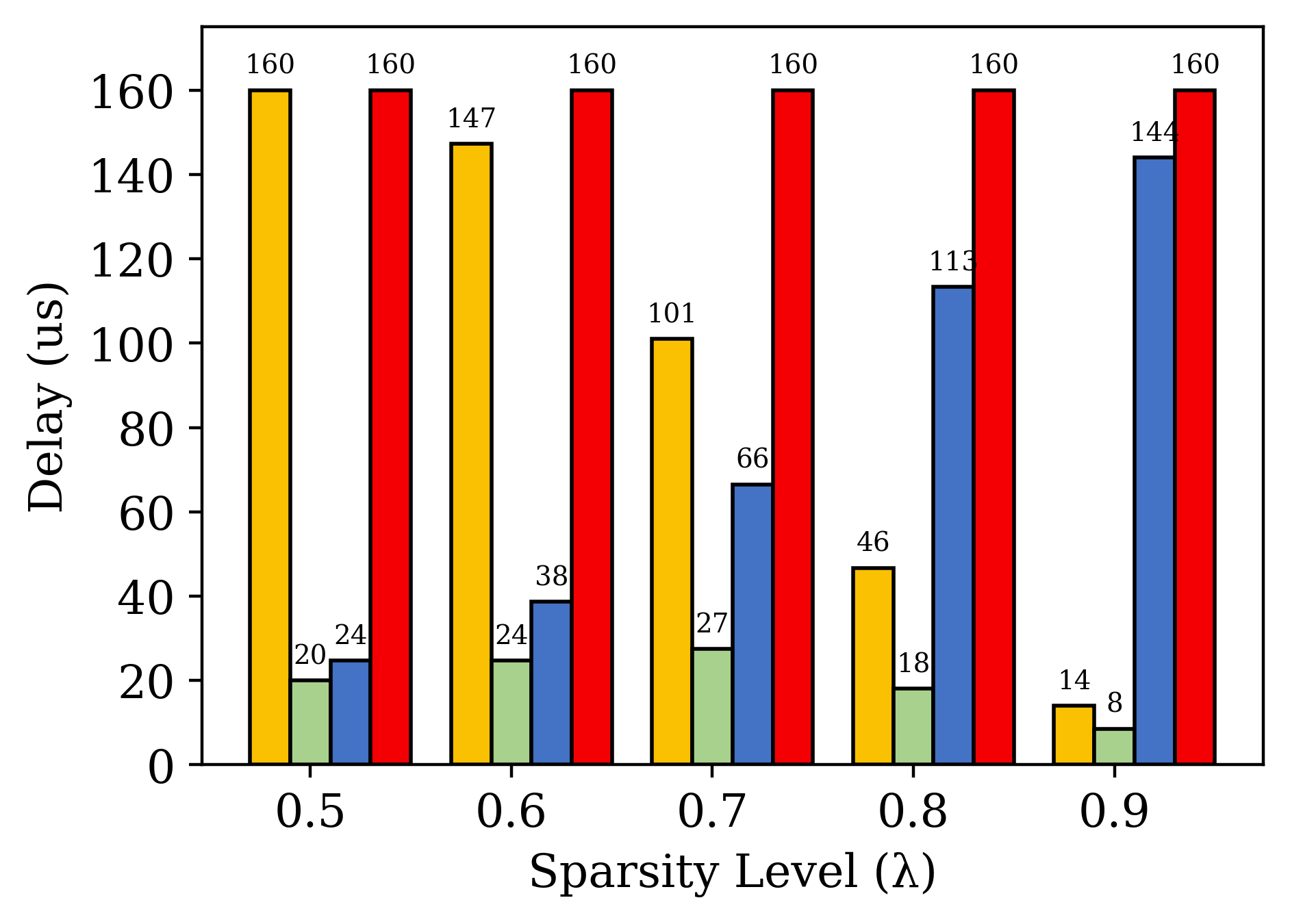}
        \caption{}
        \label{subfig:delay_comparison}
      \end{subfigure}
      \begin{subfigure}[t]{0.34\linewidth}
            \centering
        \includegraphics[width=\linewidth]{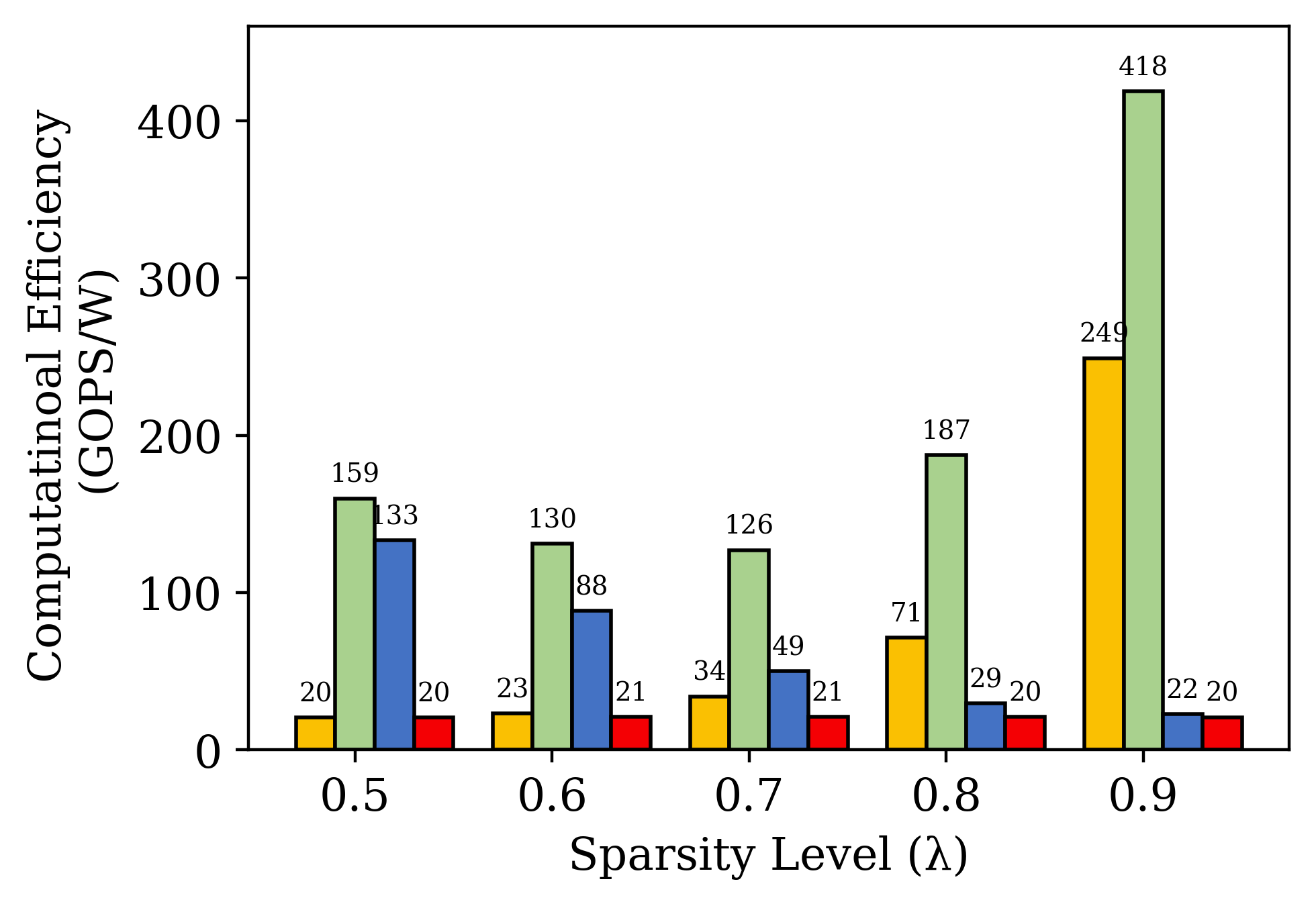}
        \caption{}
        \label{subfig:computational_efficiency_comparison}
      \end{subfigure}
    \end{minipage}

    \begin{minipage}{\textwidth}%
      \begin{subfigure}[t]{0.33\linewidth}
            \centering
        \includegraphics[width=\linewidth]{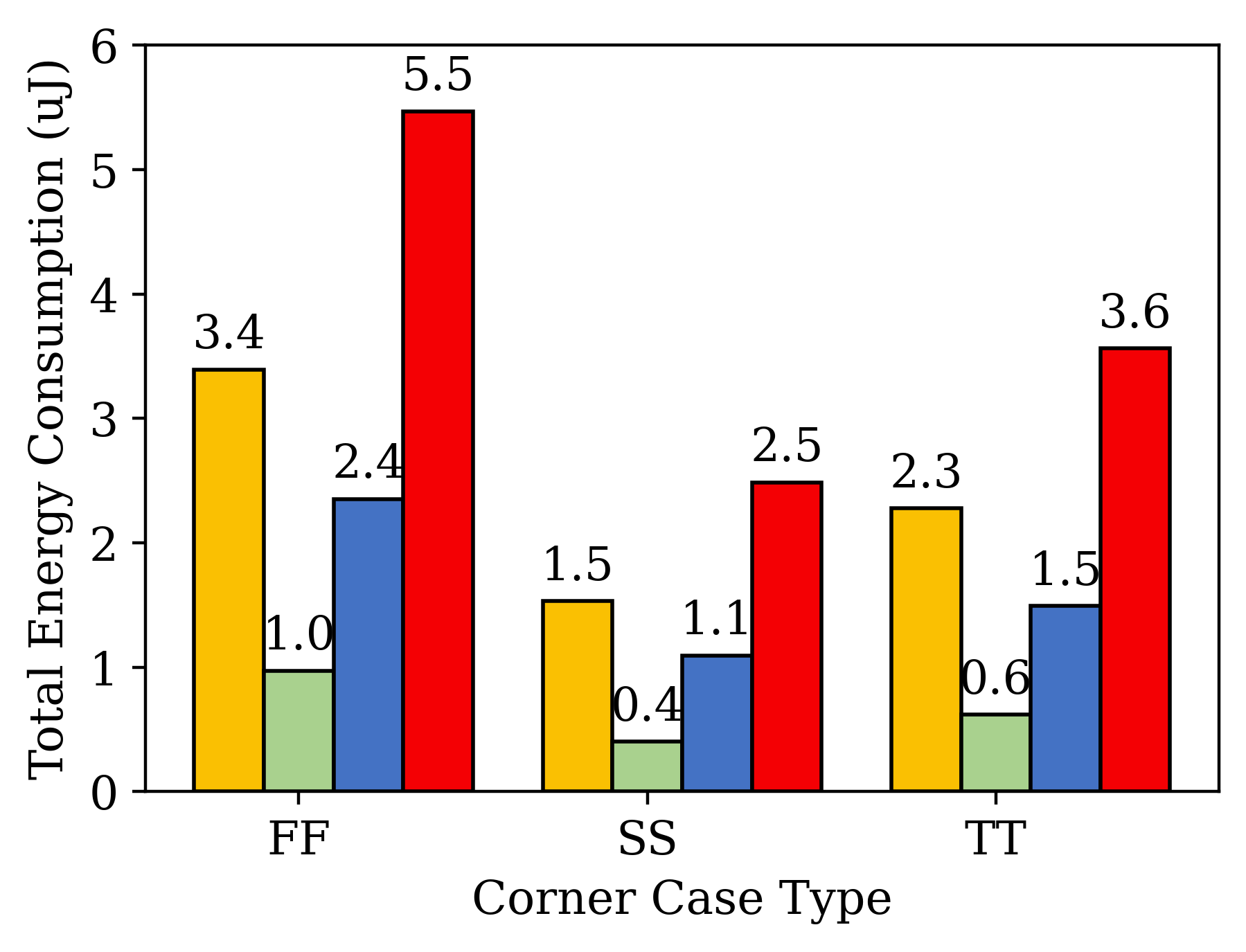}
        \caption{}
        \label{subfig:corner_analysis}
      \end{subfigure}
      \begin{subfigure}[t]{0.33\linewidth}
            \centering
        \includegraphics[width=\linewidth]{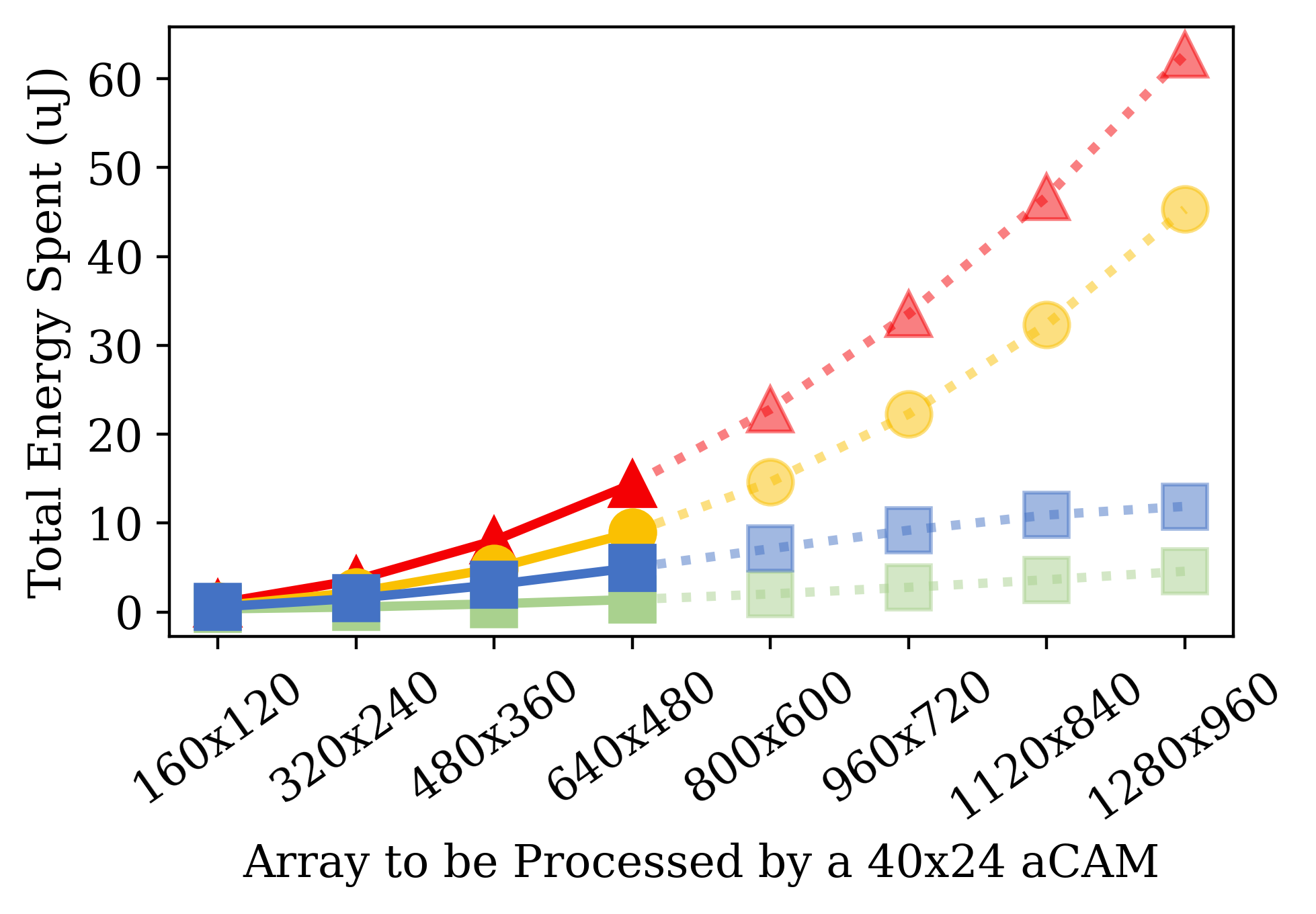}
        \caption{}
        \label{subfig:scalableCAM}
      \end{subfigure}
      \begin{subfigure}[t]{0.34\linewidth}
            \centering
        \includegraphics[width=\linewidth]{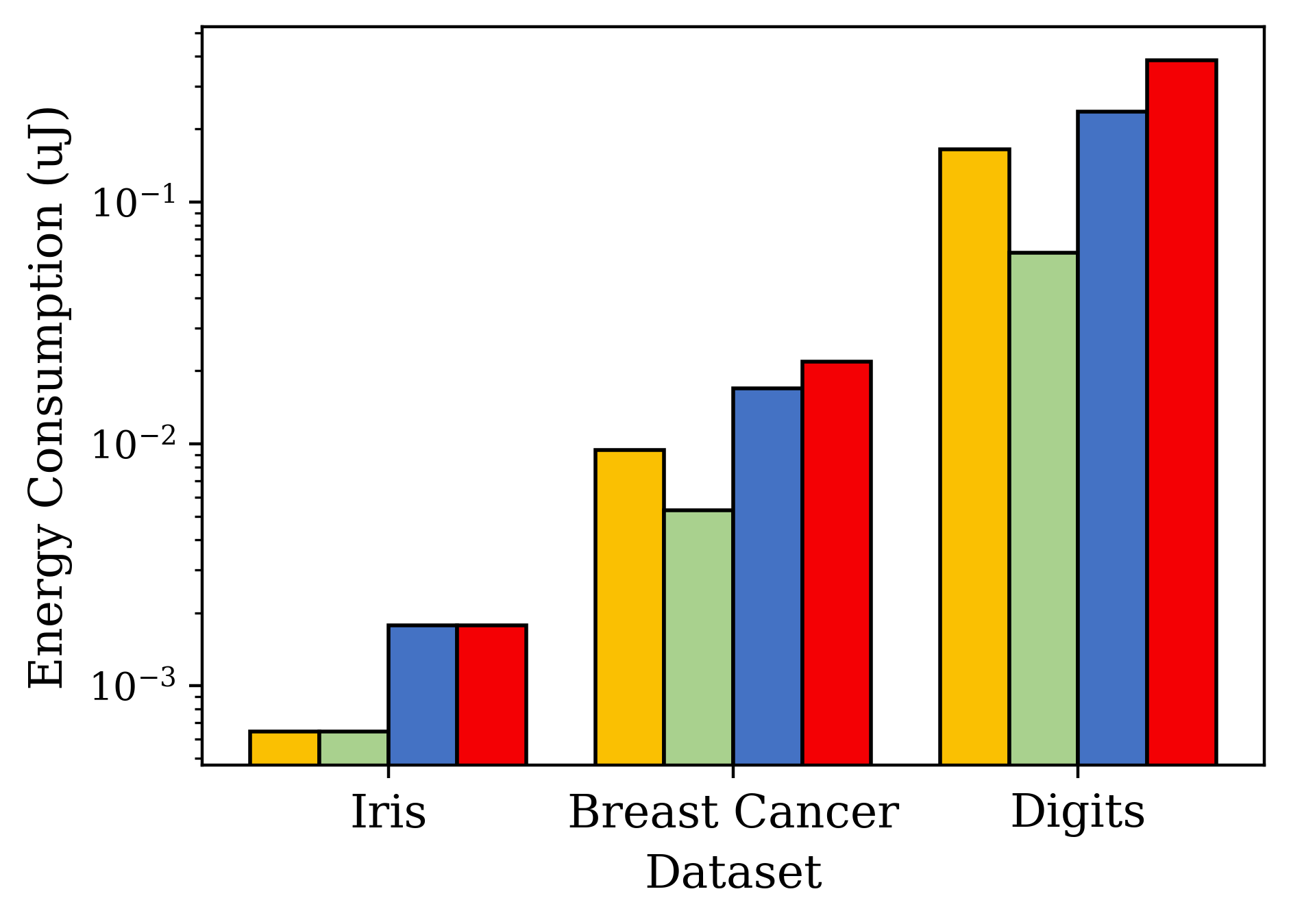}
        \caption{}
        \label{subfig:real_models}
      \end{subfigure}
    \end{minipage}

  \end{minipage}
  \caption{Comparison of (a) total energy consumption, (b) delay, and (c) computational efficiency between different optimization techniques when processing a 240$\times$320 array with a 24$\times$48 CAM. (d) Energy consumption comparison for processing a 240x320 array with a 24$\times$48 CAM at $\lambda$=0.7 across different tech nodes: typical (tt), slow NMOS and PMOS (ss), and fast NMOS and PMOS (ff). (e) Energy consumption projection: dotted lines show the predicted curve based on four actual data points connected by solid lines, with all fitted curves having an ${R}^2$ value above 97\%. (f) Cost of accelerating decision-trees trained on datasets.
  } 
  \label{subfig:experiments123}
\end{figure*}

We investigated the relationship between sparsity and structural balance in TBML models using a binary classification dataset with 1,000 samples and 5 features. Models XGBoost and RandomForest were trained with a single decision tree of max depth 10, and structural balance was measured by the node count difference between left and right subtrees from the root node. As shown in Fig.\ref{subfig:trend_in_DT}, XGBoost has a Pearson correlation of -0.35 (p-value 1.433E-148) and RandomForest -0.327 (p-value 2.809E-125) for tree balance and sparsity, demonstrating more balanced trees tend to be sparser when represented as arrays \cite{balanced_tree}. Above all, efficient processing in aCAM requires optimization techniques for variable tree structures. MonoSparse-CAM has considerable computational savings even at low sparsity where the traditional FR starts to suffer. More precisely, MonoSparse-CAM discards predictable matchline discharging at lower sparsity, while FR's selective loading is favorable for higher sparsity levels, as shown in Figs.\ref{subfig:energy_consumption_comparison} and \ref{subfig:delay_comparison}. This adaptability allows MonoSparse-CAM to achieve the gain in efficiency over a wide range of sparsity.

\section{Experiment and Results}
We evaluate the efficiency of MonoSparse-CAM based on its energy consumption, latency, and computational efficiency with respect to existing approaches. In addition, we investigate the impacts of process variations, scalability, and energy efficiency. Finally, we validate the optimization method using real datasets.


\subsection{Experimental Setup}



The analog CAM cells of 6T2M (shown in Fig.\ref{subfig:cam_cell}) were used for experiments, and energy measurements were simulated using HSPICE. A custom Python script varies the dimensions of CAMs, configures the conductances of each cell, and sets input voltages, using an 11 pF capacitor on each match line. The simulations use CMOS device models from the 65nm technology node with a specialized function that linked voltage ranges to conductances for precise control.

A Python simulator was developed to assess latency, throughput, and energy efficiency during tree-based model inference, using benchmark datasets. The simulations, validated against experimental measurements, included the analog CAM array, peripheral circuitry (sense amplifiers, pre-charge), and energy consumption for registers (calculated using CACTI). The analog CAM arrays dominated power consumption, with peripheral components contributing minimally.

Each memristor in this CAM array stored a value between 0 and 1, with 1/256 precision (0.00392 per step) allowing 256 distinct levels. In other words, such precision provided a good balance between accuracy and hardware efficiency; increasing precision further would add circuitry complexity and raise energy consumption, but the chosen precision struck an optimal balance. Note that the energy required for reprogramming the memristors was not included in our calculations. In our simulations, we assume ideal device behavior to demonstrate the system's functionality under optimal conditions. We acknowledge that real-world memristor variability may impact performance, which will be explored in future work.

\subsection{Energy Consumption and Delay Analysis}
We processed a 240$\times$320 array using a 24$\times$48 CAM configuration. Results in Fig.~\ref{subfig:energy_consumption_comparison} and Fig.~\ref{subfig:delay_comparison} show that for raw processing, total energy consumption is invariant across a wide range of sparsity values (mean = 3.77$\mu$J, stdev = 0.15$\mu$J). Data in Fig.~\ref{subfig:experiments123} are averaged over at least three trials. At high sparsity ($\lambda$ = 0.9), FR achieved 15$\times$ energy gain. However its energy efficiency decreased with lower sparsity, approaching that of raw processing. In contrast, MonoSparse-CAM showed the least energy usage at all sparsity levels (mean = 0.40$\mu$J, stdev = 0.48$\mu$J). Energy gain peaked at 22.16$\times$ over raw processing and 10.96$\times$ over FR. Similarly, latency trends follow this result as shown in Fig.~\ref{subfig:delay_comparison}.

The examination further incorporated the energy utilized by control circuitry—such as pre-charging, sense amplifiers, and register functionalities during both writing and reading processes. This encompasses the energy expenditure from a 120-byte cache that retains match line states (MLs). These elements were critical for precisely determining the overall energy consumption associated with CAM operations.

\begin{equation}\label{eq:1}
    \text{GOPS/W} = \frac{H_{LargeArray} \cdot W_{LargeArray} / \text{Delay}_{Technique}}{\text{AvgPower}_{Technique}}
\end{equation}

\subsection{Computational Efficiency and Throughput Analysis}
Fig. \ref{subfig:computational_efficiency_comparison} shows the efficiency comparison of different methods for various sparsity levels. The MonoSparse-CAM consistently exhibits the best efficiency, reaching 418 GOPS/W at a high sparsity level ($\lambda$ = 0.9), while FR presents only a medium level of efficiency. Computational efficiency is computed as shown in Eq.\ref{eq:1}. Contrasted to this, the raw and monotonicity-only processing both show low efficiencies. These results indicate tremendous performance gains with MonoSparse-CAM in bettering computational efficiency across a wide range of sparsity levels.

This is because, although a reduction in both width and height improves efficiency due to the earlier detection of mismatches and avoidance of redundant computation, thereby leading to significant operational savings, narrower width arrays are much more effective. CAM arrays are processed in tiles, evaluating matchlines based on overall tile results. Early or late mismatches within a tile don't impact processing, but narrower tiles enable earlier mismatch detection, improving efficiency. Of all the techniques, raw processing showed the lowest throughput with the highest energy consumption, while FR and Monotonicity-only methods had only a moderate improvement. MonoSparse-CAM outperformed all methods in terms of higher throughput and lower energy consumption across all sparsity levels.

\subsection{Corner and Scalability Analysis}
We performed corner analysis and scalability analysis for all CAM designs. In corner analysis, the energy consumptions under typical (TT), fast NMOS and fast PMOS (FF), and slow NMOS and slow PMOS (SS) are considered. As it can be seen from Fig.~\ref{subfig:corner_analysis}, MonoSparse-CAM consistently had the lowest energy among all corner cases and at least 3.7$\times$ lower than state-of-the-art.

In the scalability analysis, we processed arrays of increasing size (160$\times$120, 320$\times$240, 480$\times$360, 640$\times$480) using a 40$\times$24 analog CAM at $\lambda = 0.7$ sparsity. Energy usage grew exponentially for raw processing and FR, while MonoSparse-CAM exhibited approximately linear growth, as evidenced by the prediction curves (${R}^2 > 0.97$) in Fig.~\ref{subfig:scalableCAM}.

\subsection{Acceleration of Trained Decision-Tree Models}

Decision-tree models were trained, each with full processing of a different acceleration on the Iris, Breast Cancer, and Digits datasets, for real-world practical task validation. The Iris model uses a 2$\times$2 CAM tile and achieves 100\% accuracy with 95\% empty cells ($\lambda$=0.82) on a 4×10 array. A 5$\times$4 tile is used by the Breast Cancer model, which achieves 94.15\% accuracy on a 30$\times$16 array with 89.17\% of the cells empty ($\lambda$=0.73). The Digits model (84.26\% accuracy, 86.47\% empty cells or $\lambda$=0.70, 64$\times$135 array) was processed with a 8$\times$16 tile. Models were split 70/30 for training and testing, with feature scaling applied to the training set. Results are shown in Fig.~\ref{subfig:real_models}.




\section{Conclusion}
In this paper, we proposed MonoSparse-CAM, a co-designed hardware-software technique, which leverages not only the sparsity of tree-based machine learning models but also the monotonicity of CAM arrays. This approach significantly improves the processing efficiency and reduces the energy consumption: energy reduction by 28.56$\times$ over raw processing and 18.51$\times$ over state-of-the-art techniques. These represent a big step toward energy-efficient machine learning computation on hardware. Future work will consider further optimizations for large-scale applications, while the method will also be extended to other model architectures for more sustainable AI systems.


\clearpage






\end{document}